# Robots and COVID-19: Challenges in integrating robots for collaborative automation

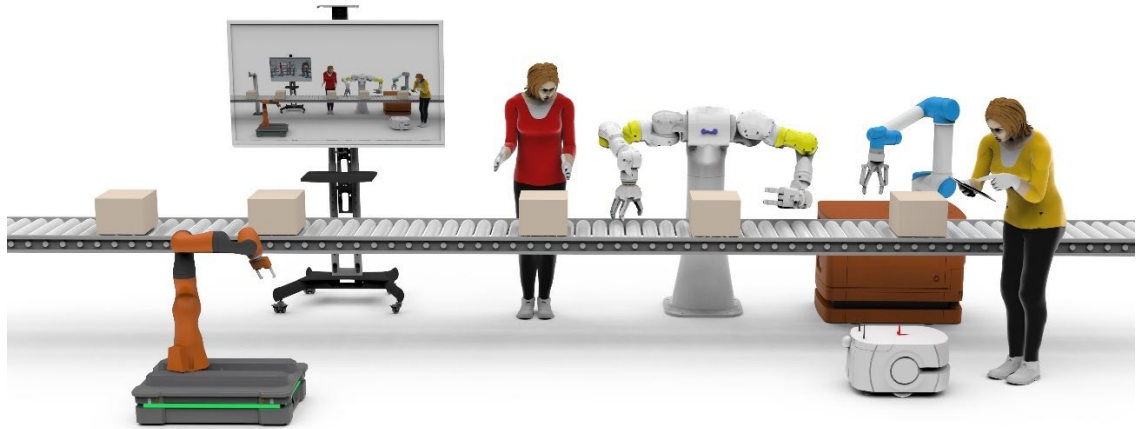

*Figure 1: An illustration of humans and robots working in close proximity at a factory floor.*

ALI AHMAD MALIK
Mads Clausen Institute
University of Southern Denmark, Denmark
Email: alimalik@mci.sdu.dk

## 1. Abstract

**Objective:** The status of human-robot collaboration for assembly applications is reviewed and key current challenges for research community and practitioners are presented.

**Background:** As the pandemic of COVID-19 started to surface the manufacturers went under pressure to address demand challenges. Social distancing measure made fewer people available to work. In such situations, robots were pointed at to support humans to address shortage in supply. An important activity where humans are needed in a manufacturing value chain is assembly. HRC assembly systems are supposed to safeguard coexisting humans, perform a range of actions and often need to be reconfigured to handle product variety. This requires them to be resilient and adaptable to various configurations during their operational life. Besides the potential advantages of using robots the challenges of using them in industrial assembly are enormous.

**Methods:** This mini review summarizes the challenges of industrial deployment of collaborative robots for assembly application.

**Applications:** The documented challenges highlight the future research directions in human-robot interaction for industrial applications.

**Keywords:** Robot; Human-robot collaboration; Assembly; Challenges; COVID-19

## 2. Introduction

The interaction of humans and robots has got ample attraction in the past years. Manufacturers have implemented robots for a range of simple to complex tasks in close proximity to humans (Wojtynek et al. 2019)(Wang et al. 2020). When robots get into action in close proximity to humans, they are referred to as collaborative robots (cobots) (Wang et al. 2019). The name of 'collaborative robots' is given to a class of robotics when compared to traditional industrial robots. Since the traditional industrial robots are operated in close fenced areas (Sheridan 2016), the elimination of fence inherently displays a degree of interaction with humans. Although, the closeness, or the level of engagement between humans and robots has been an interesting topic (Kolbeinsson et al. 2019) (Michaelis et al. 2020), and can also refer to level of automation (Malik & Bilberg 2019b) but is out of scope of this study.

For the year 2018, the increase in cobots installation worldwide was 23% as compared to 5% for the traditional industrial robots (Anon n.d.). However, the industrial application in complex tasks is very limited. The COVID-19 pandemic and the shortening of ventilators pointed towards cobots as a solution to ramp-up production volume while maintaining social distancing (Malik et al. 2020). But the desired objectives were higher than the state of industry ready technologies.

From a practitioner point of view, this article presents a review of key challenges being faced by manufacturers to integrate cobots and automate a range of manual tasks.

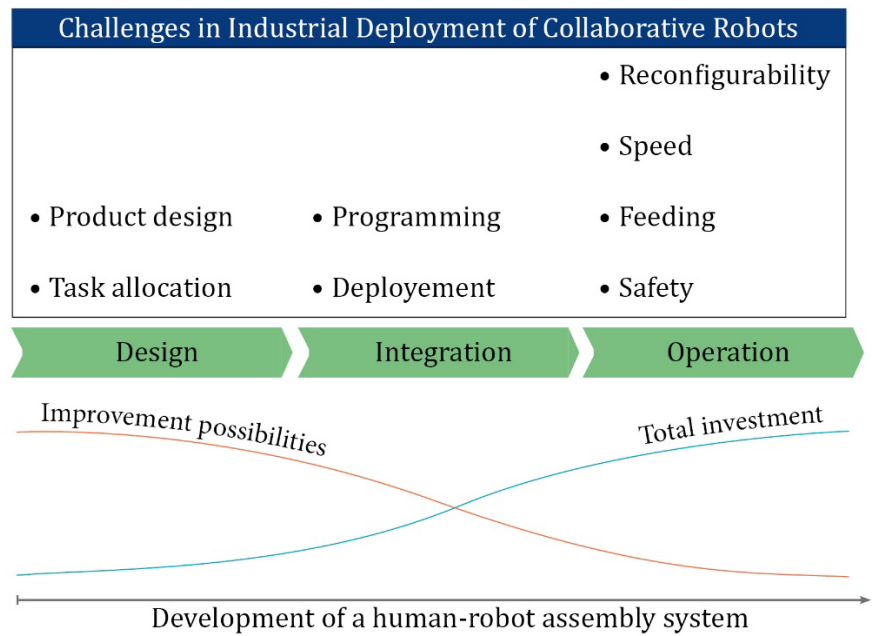

*Figure 2: Challenges in development and operation of HRC assembly systems.*

## 3. Challenges in industrial human-robot interaction

### 3.1. Slow speed of cobots

With current safety technologies and industrial standards, cobots are required to be operated at slow speed (i.e. <250mm/s) (Marvel & Norcross 2017). It is way less than speed of the human arm which can go up to 1500 mm/s (Nof 1999). If the robot is sharing the assembly workload, it has to take care of the cycle time of the preceding resource.

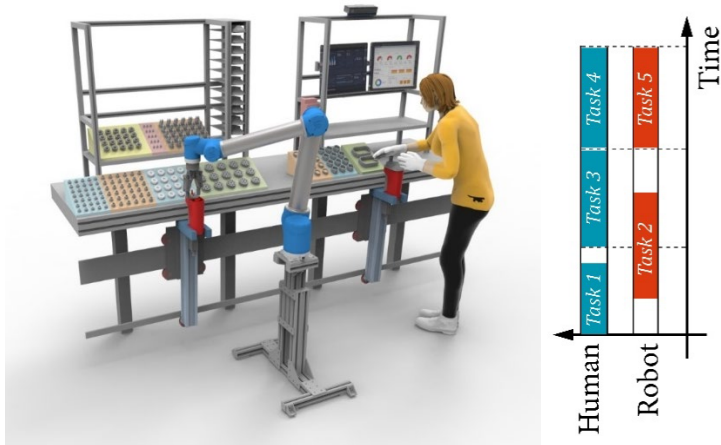

*Figure 3: An illustration of humans and robots working in close proximity at a factory floor.*

Assembly of a product often follows a sequence of actions called assembly precedence. When collaborating with another production resource (e.g. robot or human), the sequential processes must be balanced in time to take care of the preceding and subsequent tasks to avoid any bottle necks. The long cycle time of cobots makes it difficult for them to actually act as a co-worker. If the cobots are expected to be used in closer collaboration with humans, they need to attain the same speed as humans. But this requires that cobots are able to anticipate the movement and action of their collaborating human(s) and stop or change their path of motion if a collision is likely to happen.

### 3.2. Dynamics of task allocation

The complexity of a number of assembly tasks keep humans a need. Nevertheless, many of the tasks are repetitive with low complexity (Samy & ElMaraghy 2012). If identified rightly, tasks requiring human skills (sensing, dexterity or quick decision making) can be separated from less ergonomic, repetitive tasks and can be assigned to a robot working next to a human operator (Malik & Bilberg 2019a). Assigning right tasks to the right resource considering their skills and competencies is the key to achieve desired objectives of human-robot collaboration.

This problem is two folded. One aspect of task assignment to the human and robot is needed when planning an automation process. However, during the operation, the robot must be able to take or relive tasks as the need arises (Bilberg & Malik 2019). In the first scenario, the industrial practice is to select low complexity tasks by gut-feeling. The alternatives are complex mathematical models that are impractical for industrial applications. The second scenario is more challenging where dynamic task allocation is desired. (Tsalatsanis et al. 2012)(Müller et al. 2017). The Fitts list of human-machine interaction is well obsolete now (Sheridan 2016) and there is a need to have models describing human factors and suitability of humans and robots for any given task.

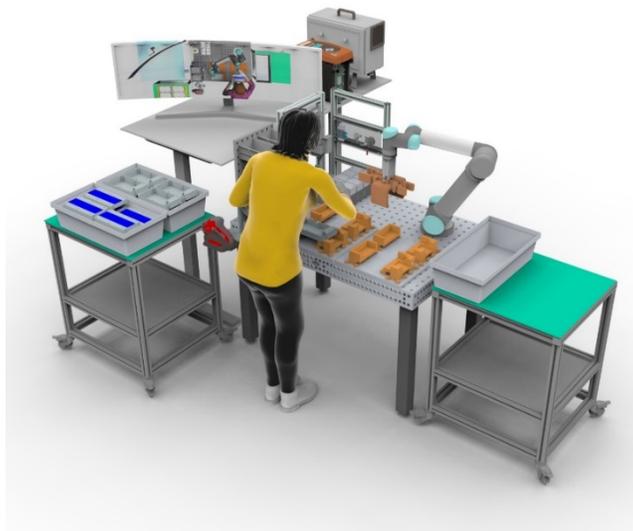

*Figure 4: Dynamic task allocation is needed when humans and robots share assembly tasks workload.*

Nevertheless, the identification and assignment of low complexity tasks to the robot must be an automated activity starting from the design of the product as metadata and must be 'built-in' into the design of a product just like a bill of material (BOM). A structured quantification of tasks for ease of automation would make it possible to prioritize the tasks based on their automation potential in an assembly process thereby simplifying the work-load distribution. The quantification will also help in formulating logics for a dynamic online HRC workload balancing.

### 3.3. Safety of human operators

A robot must not cause an injury to a human or, through inaction, allow a human being to come to harm (Boden et al. 2017). The statement comes from fiction and is the first law of robotics as described by Asimov (Asimov 2004). For industrial applications, when humans and robots get into a physical contact every possibility of an accident or injury needs to be evaluated (Pedrocchi et al. 2013) (Hentout et al. 2019). A human-robot collaboration must not cause an injury to any human, either directly or indirectly, during an operation or during a failure. Measures for safe design of robot's mechanical structure as well as for collaborative applications have been defined in international standards such as ANSI/RIA 2012; ISO/TS 15066 2016 (Vemula et al. 2018).

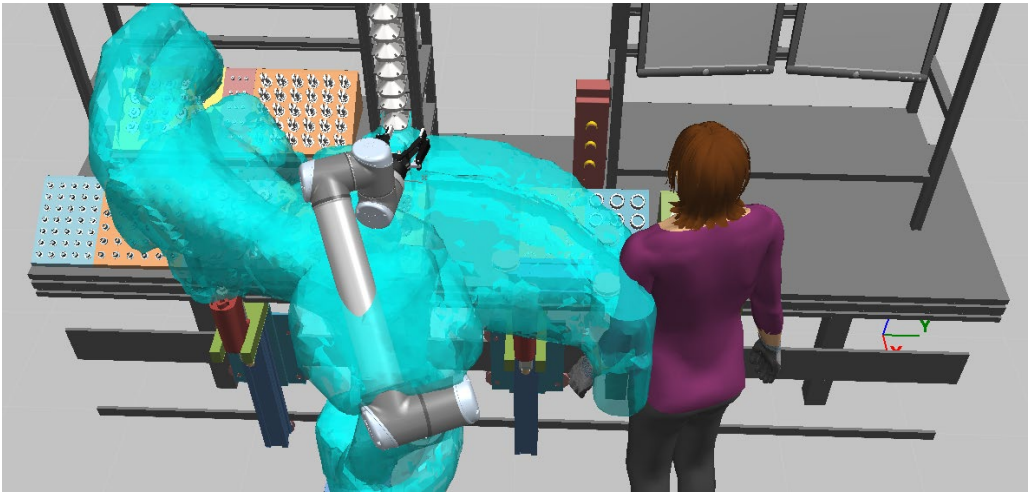

*Figure 5: Robot work envelope in a human-robot shared workspace.*

These safety standards for collaborative robots enforce severe restrictions to robot motion (Zanchettin et al. 2016) resulting into conservative speed limits (<250mm/sec). A way to answer this is by implementing speed and separation monitoring that requires the robot to reduce its speed as the distance between human and robot gets smaller. Several techniques of ensuring safety to humans have been defined in the literature such as safety eye, tracking of human movement, active collision avoidance etc. (Wang et al. 2019). However, the maximum allowable speed limit and getting a system safety validated are the challenges on the way to integrate cobots in assembly applications.

### 3.4. Deployment time

From author's theorization and other studies (Bauer et al. 2016), it has been observed that the integration of cobots often takes longer time than what is anticipated at the planning stage. This becomes even longer for complex tasks such as assembly. The unpredicted undesirables of the planning stage that appear later in the project are in relation to process balancing, feeding, fixtures, need of multiple grippers and safety complications. It is due to the fact that robots do not perform in isolation. Their movements need to be coordinated with other hardware (machine tools, equipment., end of arm tooling), peripherals (vision systems, force sensors) as well as humans (Anon 2019). The coordination of all of this gives rise to the integration and operational complexity.

Several methods and frameworks have been documented for developing human-robot hybrid production systems (Tan et al. 2009) (Francalanza et al. 2014) (Malik & Bilberg 2017). However, new approaches are required aimed at minimizing the time and effort for integration, validation and reconfiguration (Malik & Bilberg 2018). Ready to deploy hardware, software templates for programming, easier programming interfaces and computer simulations can make possible the quick deployment of cobots.

### 3.5. Teaching and programming

Contrary to conventional industrial robots, the programming of cobots in assembly is not a one-time activity in their operational life. The needed flexibility and adaptability require easy ways of programming the cobots.

The most common approach to teach a robot a manipulation task is by using the programming language that comes with the robot, often through a teach pendant (online programming). It is the most reliable method for assembly tasks where precise adjustments are needed. However, these languages are proprietary and unique to each robot brand. As for now there are more than 30 robot programming languages (Anon 2019). Another approach is using offline graphical or non-graphical simulation tools, called offline programming . However, these techniques are still time consuming for frequent changes.

The new robots are mechanically compliant, enabling a programmer to move robot's hand to teach a manipulation task (Sheridan 2016). But is not suitable for tasks requiring precise adjustments because the accuracy of a task is then dependent on the human that guides the robot. Another approach is observing human subjects in performing manipulation tasks (Shah et al. 2011). By using markers on the humans limbs, the observation is seen by computer vision. The use of augmented reality (AR) has been demonstrated to teach a robot a task. However, easier, and faster methods for programming are needed for fast integration and reconfiguration. It must be as easy as using a smart phone.

### 3.6. Reconfiguration

The need of variety in manufacturing is growing. Assembly is identified as an area with highest potential to introduce product variety (ElMaraghy 2014). Batch size one is a vision for the manufactures at the verge of the fourth industrial revolution. However, cobots are not flexible enough to be reconfigured at this pace. Robots must be able to have features, hardware and software solutions that make them easy to adapt to changes, and variations in product design.

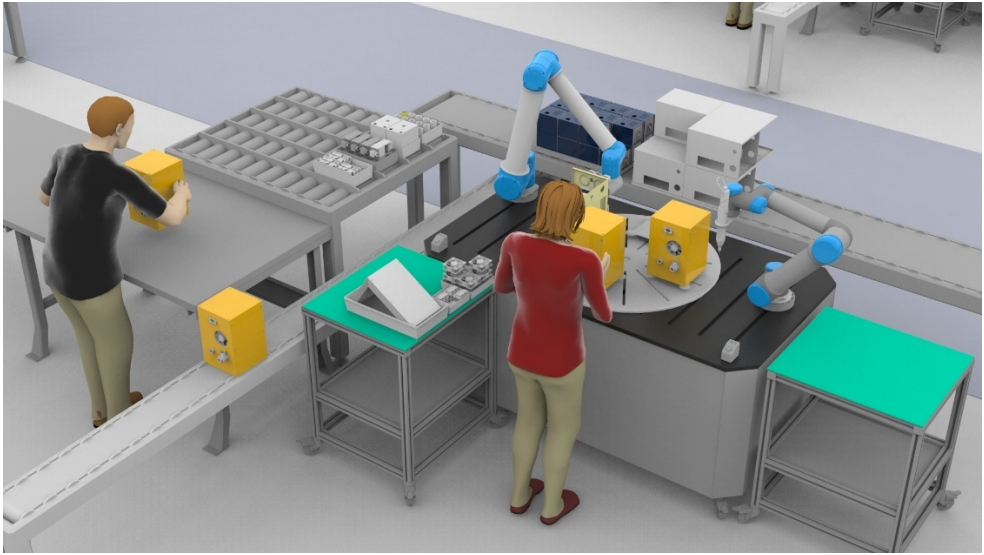

*Figure 6: Reconfigurability in robotic tasks is needed to manage variety.*

### 3.7. Product design for cobot assembly

Most cobot applications are targeted to automate assembly of existing products. These products are often designed taking care of design for assembly guidelines (Boothroyd 1994). Design for manufacturing (DFM) and design for assembly (DFA) are set of techniques that encourage the products to be designed for ease and cost effective manufacturing and assembly (Bogue 2012). However, these guidelines are aimed either for manual assembly, robotic assembly or automated assembly. It is also recognized that the design rules of one category affect differently to the other category. Furthermore, Owen (Owen 1985) argued that a product designed for manual assembly must not be produced with automated assembly and must undergo design changes before it is automated.

DFA techniques for robotic assembly are available in the literature but they don't acknowledge the aspects of a human-robot collaborative environment e.g. collaborative environments prefer lightweight robots with low payload capacity for HRC environment. A mix of rules for design for manual and robotic assembly are needed to define product design rules for ease of HRC assembly.

### 3.8. Presentation and feeding of assembly parts

To perform an assembly task, the assembly components need to be fed into the assembly station and the robot needs to be aware of the location and pose of the component to be picked (Malik et al. 2019). Feeding has two primary components i.e. part structuring and part presentation (Hansson et al. 2017). Part structuring is singulation of individual parts from a bulk and arranging it to have a known orientation, while part presentation defines moving an individual structured part to the point of use or assembly.

Conventional part feeding techniques utilize mechanical structures e.g. vibratory bowl feeders. Mechanical feeders are specific to a single part geometry (Edmondson & Redford 2001) and are inflexible to any variation in component's shape. The focus has shifted toward using machine vision to develop flexible feeding mechanisms. The challenges are to have feeding systems that are easy to deploy, easy to integrate and are cost effective.

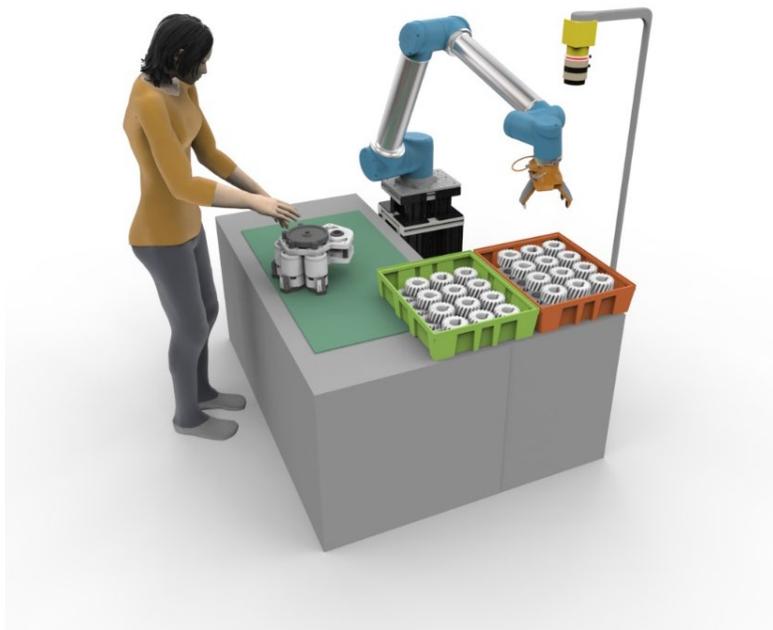

*Figure 7: Presentation of parts to the robot for assembly tasks.*

### Conclusion

After all the years of automation humans are (still) essential for manufacturing arena and they will continue to be. Collaborative robots or cobots have emerged to offer a huge potential for increasing the level of automation and be human colleagues. They can be useful to develop factories that are more resistant to pandemics. Several laboratory solutions have been developed in the past years to pave the path for robots as teammates in factories but still a lot needs to be done. This mini review points to a list of challenges that manufacturer are facing on the way to integrate cobots as teammates. The summary is of value to the researchers to set research directions and for companies that are aiming to implement cobots at their production floor.

**References**

Asimov, I., 2004. I, Robot [1950]. *Bantam Dell, New York*.

Bauer, W. et al., 2016. Lightweight robots in manual assembly—best to start simply!: Examining companies' initial experiences with lightweight robots,. *Frauenhofer-Institut für Arbeitswirtschaft und Organisation IAO, Stuttgart*.

Bilberg, A. & Malik, A.A., 2019. Digital twin driven human-robot collaborative assembly. *CIRP Annals - Manufacturing Technology*, 68(1), pp.499–502.

Boden, M. et al., 2017. Principles of robotics: regulating robots in the real world. *Connection Science*, 29(2), pp.124–129.

Bogue, R., 2012. Design for manufacture and assembly: background, capabilities and applications. *Assembly Automation*, 32(2), pp.112–118.

Boothroyd, G., 1994. Product design for manufacture and assembly. *Computer-Aided Design*, 26(7), pp.505–520.

Edmondson, N. & Redford, A., 2001. Flexible parts feeding for flexible assembly. *International Journal of Production Research*, 39(11), pp.2279–2294.

ElMaraghy, H., 2014. Managing variety in manufacturing. *Procedia CIRP*, (17), pp.1–2.

Forbes, Demand for collaborative robots continues to grow. Available at: https://www.forbes.com/sites/michaelmandel1/2019/11/01/demand-for-collaborative-robots-continues-to-grow/#322780ca4cb9 [Accessed May 31, 2020].

Francalanza, E., Borg, J. & Constantinescu, C., 2014. Deriving a systematic approach to changeable manufacturing system design. *Procedia CIRP*, 17, pp.166–171.

Hansson, M.N. et al., 2017. Modelling Capabilities for Functional Configuration of Part Feeding Equipment. *Procedia Manufacturing*, 11, pp.2051–2060.

Hentout, A. et al., 2019. Human-robot interaction in industrial collaborative robotics: a literature review of the decade 2008-2017. *Advanced Robotics*, 33(15-16), pp.764–799.

Kolbeinsson, A., Lagerstedt, E. & Lindblom, J., 2019. Foundation for a classification of collaboration levels for human-robot cooperation in manufacturing. *Production & Manufacturing Research*, 7(1), pp.448–471.

Malik, A.A., Andersen, M.V. & Bilberg, A., 2019. Advances in machine vision for flexible feeding of assembly parts. *Procedia Manufacturing*, 38, pp.1228–1235.

Malik, A.A. & Bilberg, A., 2019a. Complexity-based task allocation in human-robot collaborative assembly. *Industrial Robot: the international journal of robotics research and application*, 46(4), p. pp. 471–480.

Malik, A.A. & Bilberg, A., 2019b. Developing a reference model for human-robot interaction. *International Journal on Interactive Design and Manufacturing (IJIDeM)*, pp.1–7.

Malik, A.A. & Bilberg, A., 2018. Digital twins of human robot collaboration in a production setting. *Procedia Manufacturing*, 17, pp.278–285.

Malik, A.A. & Bilberg, A., 2017. Framework to implement collaborative robots in manual assembly: a lean automation approach. In *Proceedings of the 28th DAAAM International Symposium, B. Katalinic (Ed.), Published by DAAAM International, ISSN*. pp. 1726–9679.

Malik, A.A., Masood, T. & Kousar, R., 2020. Reconfiguring and ramping-up ventilator production in the face of COVID-19: Can robots help? *arXiv preprint arXiv:2004.07360*.

Marvel, J.A. & Norcross, R., 2017. Implementing speed and separation monitoring in collaborative robot workcells. *Robotics and computer-integrated manufacturing*, 44, pp.144–155.

Michaelis, J.E. et al., 2020. Collaborative or Simply Uncaged? Understanding Human-Cobot Interactions in Automation. In *Proceedings of the 2020 CHI Conference on Human Factors in Computing Systems.* pp. 1–12.

Müller, R., Vette, M. & Geenen, A., 2017. Skill-based Dynamic Task Allocation in Human-Robot-Cooperation with the Example of Welding Application. *Procedia Manufacturing*, 11, pp.13–21.

Nof, S.Y., 1999. *Handbook of industrial robotics*, John Wiley & Sons.

Owen, T., 1985. *Assembly with robots*, Springer Science & Business Media.

Pedrocchi, N. et al., 2013. Safe human-robot cooperation in an industrial environment. *International Journal of Advanced Robotic Systems*, 10(1), p.27.

Ready Robotics, 2019. *5 reasons programming robots is so hard*. Technical Report by Ready Robotics, U.S.A. Samy, S. & ElMaraghy, H., 2012. Complexity mapping of the product and assembly system. *Assembly automation*, 32(2), pp.135–151.

Shah, J. et al., 2011. Improved human-robot team performance using chaski, a human-inspired plan execution system. In *Proceedings of the 6th international conference on Human-robot interaction*. pp. 29–36.

Sheridan, T.B., 2016. Human-robot interaction: status and challenges. *Human factors*, 58(4), pp.525–532.

Tan, J.T.C. et al., 2009. Human-robot collaboration in cellular manufacturing: Design and development. In *Intelligent Robots and Systems, 2009. IROS 2009. IEEE/RSJ International Conference on*. pp. 29–34.

Tsalatsanis, A., Yalcin, A. & Valavanis, K.P., 2012. Dynamic task allocation in cooperative robot teams. *Robotica*, 30(5), pp.721–730.

Vemula, B., Matthias, B. & Ahmad, A., 2018. A design metric for safety assessment of industrial robot design suitable for power-and force-limited collaborative operation. *International journal of intelligent robotics and applications*, 2(2), pp.226–234.

Wang, L. et al., 2020. Overview of Human-Robot Collaboration in Manufacturing. In *Proceedings of 5th International Conference on the Industry 4.0 Model for Advanced Manufacturing*. pp. 15–58.

Wang, L. et al., 2019. Symbiotic human-robot collaborative assembly. *CIRP Annals - Manufacturing Technology*, 68(2), pp.701–726.

Wojtynek, M., Steil, J.J. & Wrede, S., 2019. Plug, Plan and Produce as Enabler for Easy Workcell Setup and Collaborative Robot Programming in Smart Factories. *KI-Künstliche Intelligenz*, 33(2), pp.151–161.

Zanchettin, A.M. et al., 2016. Safety in human-robot collaborative manufacturing environments: Metrics and control. *IEEE Transactions on Automation Science and Engineering*, 13(2), pp.882–893.